\documentclass{article}

\usepackage{arxiv}

\usepackage[utf8]{inputenc} 
\usepackage[T1]{fontenc}    
\usepackage{hyperref}       
\usepackage{url}            
\usepackage{booktabs}       
\usepackage{amsfonts}       
\usepackage{nicefrac}       
\usepackage{microtype}      
\usepackage{lipsum}
\usepackage{graphicx}
\graphicspath{{./media/}}

\title{Revising FUNSD dataset for key-value detection in document images}

\author{
  Hieu M. Vu\\
  University of Engineering and Technology\\
  Vietnam National University\\
  Hanoi, Vietnam\\
  \texttt{vmhieu17@gmail.com} \\
  Cinnamon AI\\
  \texttt{ian@cinnamon.is}\\
   \And
 Diep Thi-Ngoc Nguyen \\
  University of Engineering and Technology\\
  Vietnam National University\\
  Hanoi, Vietnam \\
  \texttt{ngocdiep@vnu.edu.vn} \\
  HbLab Inc\\
  \texttt{chupi@hblab.vn}

}

\begin{document}
\maketitle

\begin{abstract}
FUNSD is one of the limited publicly available datasets for information extraction from document images. The information in the FUNSD dataset is defined by text areas of four categories (\textit{``key'',``value'', ``header'', ``other''}, and \textit{``background''}) and connectivity between areas as key-value relations. Inspecting FUNSD, we found several inconsistency in labeling, which impeded its applicability to the key-value extraction problem. In this report, we described some labeling issues in FUNSD and the revision we made to the dataset. We also reported our implementation of for key-value detection on FUNSD using a UNet model as baseline results and an improved UNet model with Channel-Invariant Deformable Convolution.
\end{abstract}


\section{Introduction}
\label{intro}

Forms have been a convenient way to collect and show information in many businesses. There is a huge demand in digitizing those forms and extracting  important information from them. However, the formats used in forms varies from a situation to another and such wide variations raise a great challenge to automatically extract information.

Information from a form includes not only its transcription but also the key-value (or label-value) relationships between text areas in the form. Those relationships play an important role in knowing the semantic meaning of elements in the forms. As stated in \cite{davis2019deep}, the key-value relationships in forms can be purely inferred from visual perspective. In order to support computer vision community in the development of this task, there are two image datasets for key-value extraction: FUNSD~\cite{jaume2019funsd} and NAF\cite{davis2019deep}. 

The National Archives Forms (NAF) dataset contains historical form images from United States National Archives, in which the annotated data are 165 in total. The 165 images are splitted into train/validation/test as 143/11/11, respectively. The dataset is available at \url{ http://github.com/herobd/NAF_dataset}.

In the other hand, the FUNSD has 199 document images and is splitted into train/test with 149/50 ratio. Considering the quality of images and the variety of forms in each dataset, we used FUNSD for further investigation. 

In this report we will describe several annotating issues in FUNSD and the change we made to it. After that, we will describe the baseline implementation on it and our improved model for the key-value detection problem.

The revised version of the FUNSD dataset is publicly available at \url{http://shorturl.at/cowA9}.

\section{FUNSD dataset}
\label{datastatistic}
FUNSD \cite{davis2019deep} is composed of 199 document images, which is a subset sampled from the RVL-CDIP dataset \cite{harley2015evaluation}. The RVL-CDIP dataset consists of 400,000 grayscale images of various documents from the 1980s-1990s. Those images are scanned documents, which have a low resolution of 100 dpi and low quality with various types of noise added by successive scanning and printing procedures. The RVL-CDIP dataset categorized its images into 4 classes: letter, email, magazine, form.  The author of the FUNSD dataset manually checked the 25,000 images from the “form” category, discarded unreadable and duplicate images, which leave them with 3,200 images, out of which 199 images were randomly sampled for annotation to make the FUNSD dataset. Originally, the RVL-CDIP is a subset of another dataset called Truth Tobacco Industry Document\footnote{\url{https://www.industrydocuments.ucsf.edu/tobacco/}} (TTID), which is an archive collection of scientific research, marketing and advertising documents of the largest tobacco companies in the US.

The forms in the FUNSD dataset are annotated so that they can be used for many document understanding tasks such as text detection, text recognition, spatial layout understanding, and question-answer pair extraction \cite{jaume2019funsd}.

\subsection{Data statistics}
In the FUNSD dataset, the label of each image is contained in a JSON file, in which the form is represented as a list of interlinked semantic entities. Each entity consists of a group of words that belong together both semantically and spatially. Each semantic entity is associated with a unique identifier, a label (i.e., header, question, answer and other), a bounding box, a list of words belongs to said entity, and a list of links with other entities. Each word is described by its contextual content (i.e., OCR label) and its bounding box. Noted that the “question” and “answer” label is analogous to “key” and “value”; the bounding boxes are in the form of [left, top, right, bottom] and the links are formatted as a pair of entity identifiers with the identifier of the current entity being the first element.

In the 199 images, there are in total of over 30,000 word-level annotations and about 10,000 entities. The dataset was split into the training set and the testing set with 149 images in the training set and 50 images in the testing set. More detailed statistics are described in Tab.\ref{form_count} and Tab. \ref{entity_count} (all statistics are according to the FUNSD dataset paper \cite{jaume2019funsd}).

\begin{table}[]
\caption{Counts of forms, words, entities and relations in the FUNSD dataset}
\centering
\begin{tabular}{|l|l|l|l|l|}
\hline
Subset   & No. Forms & No. Words & No. Entities & No. Relations \\ \hline
Training & 149       & 22,512    & 7,411        & 4,236         \\ \hline
Testing  & 50        & 8,973     & 2,332        & 1,076         \\ \hline   
\end{tabular}
\label{form_count}
\end{table}

\begin{table}[]
\caption{Class distribution of the semantic entities in FUNSD dataset}
\centering
\begin{tabular}{|l|l|l|l|l|l|}
\hline
Subset   & Header & Question & Answer & Other & Total \\ \hline
Training & 441    & 3,266    & 2,802  & 902   & 7,411 \\ \hline
Testing  & 122    & 1,077    & 821    & 312   & 2,332 \\ \hline
\end{tabular}
\label{entity_count}
\end{table}


\subsection{Inconsistency issues}

As well mentioned by the authors of FUNSD \cite{jaume2019funsd}, the main limitation of this dataset comes from the fact that ``there is no exact definition of \textit{what} a form is or \textit{how} we should represent it''. 

\begin{figure}
\centering
\includegraphics[width=1\textwidth]{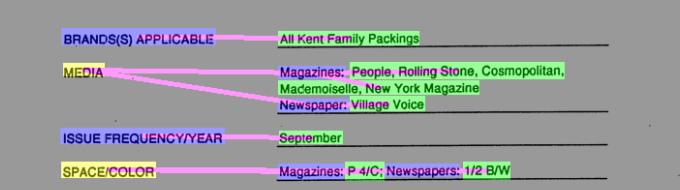}
\caption{The ``MEDIA'' and ``SPACE/COLOR'' texts are not actually headers but are just a way to represent changing relation.}
\label{fig:wrong_header}
\end{figure}

\begin{figure}
\centering
\includegraphics[width=1\textwidth]{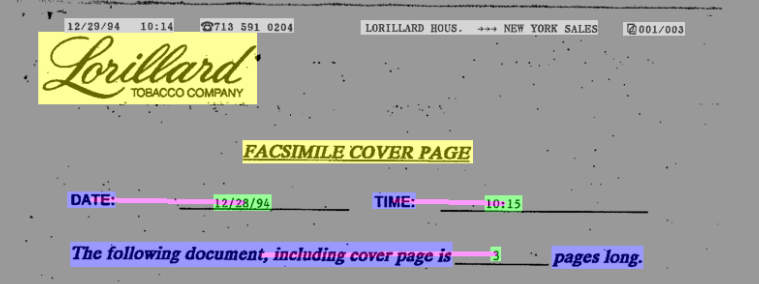}
\caption{``Lorillard'' is not a header but just a logo. The header class is used ambiguously in this case.}
\label{fig:ambiguous_header}
\end{figure}

Upon inspecting the FUNSD dataset, there are several annotating errors arise:
\begin{itemize}
    \item Incorrect textual contents (i.e. OCR label).
    
    \item Incorrect bounding boxes for words.
    
    \item Inconsistency in defining bounding boxes - an entity could be either: one big block of text, lines of text, a word, or a few consecutive words. Sometimes a word is also represented as more than one consecutive words.
    
    \item Uncanny relations: some relations that do not follow the same logic of other relations. These relations are most likely created by mistakes or typos.
    
    \item Inconsistency in annotating relations. The “header” class is sometimes used to indicates headers, sometimes as a mean to represent chaining relation (i.e., hierarchical relations that have more than 2 levels of depth) (Fig. \ref{fig:wrong_header}, and sometimes it is used ambiguously (Fig. \ref{fig:ambiguous_header}).
\end{itemize}

\section{FUNSD revision}
\begin{figure}
\centering
\includegraphics[width=1\textwidth]{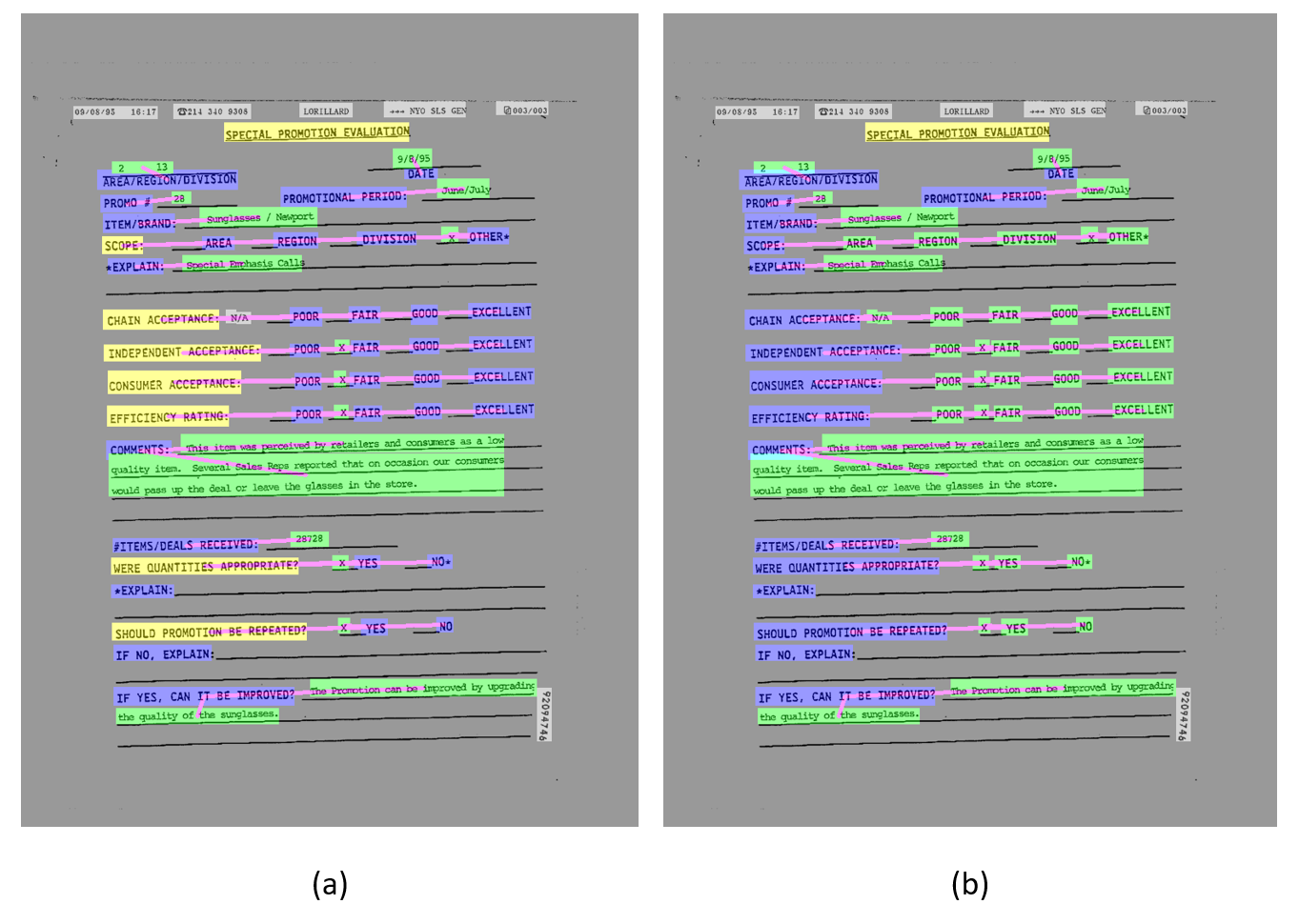}
\caption{Visualization of an image with erroneous annotations (a) and the revised annotations (b)}
\label{fig:revised}
\end{figure}

In order to use the data for developing the key-value detection network, the following steps were conducted to revise the dataset:
\begin{itemize}
    \item Visualize the annotation for all images in the dataset, an example is shown in Fig. \ref{fig:revised}a.
    
    \item Manually check each image and look for errors in annotations for relations.
    
    \item For each relation chain, only the last entity is assigned as “answer”, all other entities in the chain are assigned as “question”. Any entity than not a part of any relationship is labelled as “other”.
    
    \item During revising the relation, any incorrectness in textual content and bounding box is also addressed. However, this means that there can still be errors of these types remaining in the revised dataset.
    
    \item The result of this process is visualized again, an example is shown in \ref{fig:revised}b.
\end{itemize}

\section{A baseline of key-value detection on FUNSD}
\subsection{Layout-aware key-value detection model}
We propose an end-to-end key-value detection model directly from document images without text recognition involved.

The intuition behind this design is that: human can perform key-value pair detection even without knowing the content of the document, the contextual information is useful when performing key-value pair identification. For example, given a document written in a totally foreign language, one can easily detect groups of text that potential have key-value relation; and if asked for identify phone numbers in the document, only then the person would need to know the semantic meaning of the text. That proves that spatial layout information is enough for the human to detect key-value pairs to some extent, thus motivates the idea of combining text detection and key-value detection in an information extraction system, making the system more layout-aware.

\begin{figure}
\centering
\includegraphics[width=1\textwidth]{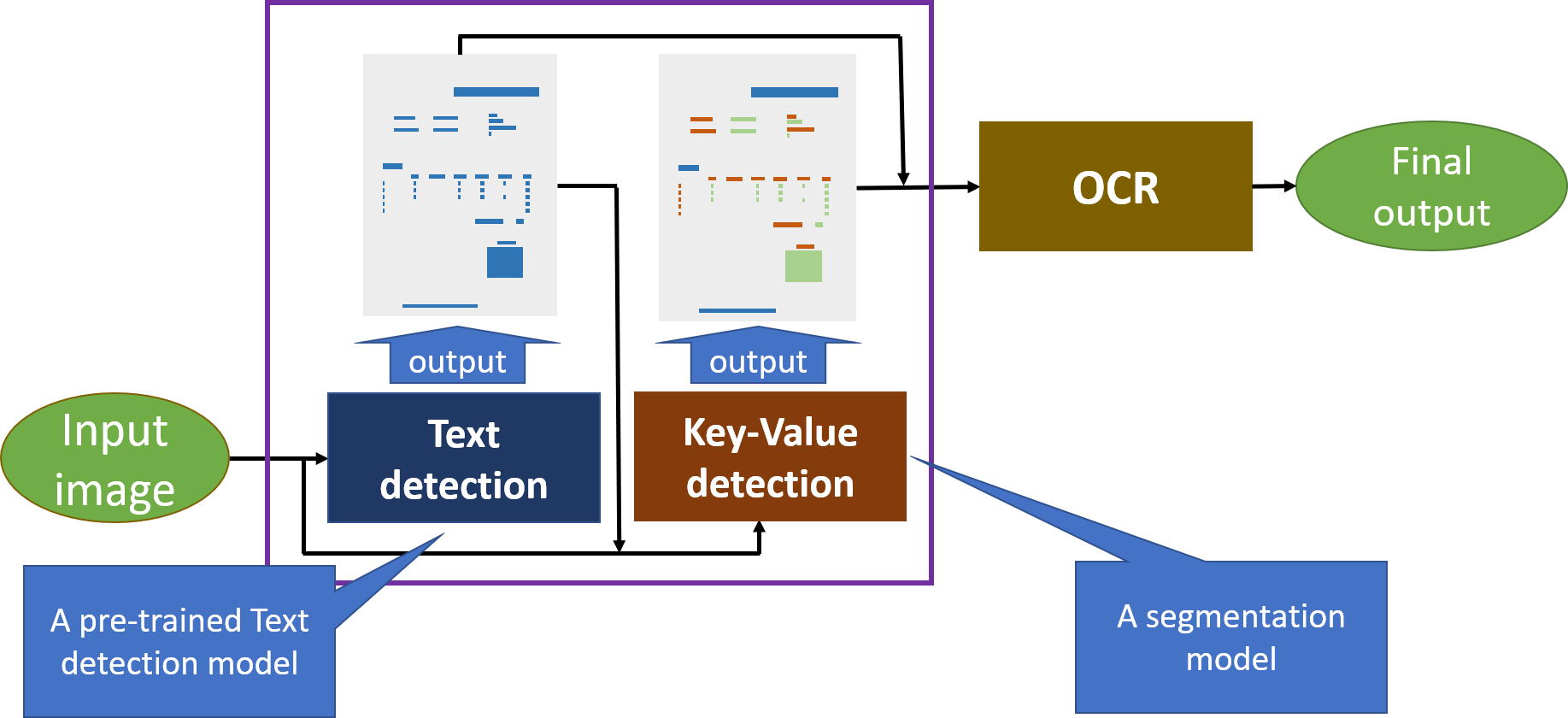}
\caption{Overall pipeline of our layout-aware key-value extraction model, where text detection and key-value detection steps are integrated into one end-to-end trainable model.}
\label{fig:pipeline}
\end{figure}

The overall pipeline of our model is shown in Fig. \ref{fig:pipeline}. The first stage is a text detection network that produces a text segmentation map. The second stage which is the key-value detection stage takes the direct output of the text mask produced by the text detection step as input. Furthermore, it additionally uses the input image as input to gain access to structural layout information of the input image, making this an end-to-end layout-aware key-value segmentation model. The second stage’s input would be the text mask and the input image concatenated together deep-wise to make a 2-channel mask, and the output is a 4-channel mask with the channels corresponding to 4 classes: key, value, other, and background. 

In this paper, we would show that using only spatial information is enough to determine key and value components in a document image. Detected key and value components can then be paired together using nearest neighbour mapping, and lastly processed by an OCR model to get the final output.

\subsection{Related works}

Much of previous works assume the availability of templates for form types of interest~\cite{barrett2004digital,butt2012information,hammami2015one}. This assumption has been relaxed in later work where a table format, cells, or other regularity can be assumed~\cite{zhou2016irmp,hirayama2011development}. 

However, many forms do not have table structures. Several approaches treat finding key-value pairs in a form as creating scene graphs from images. In those approaches, the objects in the images are nodes and edges are relationships between the objects. 

Zhang et al.~\cite{zhang2017relationship} uses a detection network to predict object, and prune the relationships based on whether the locations are spatially coherent. Yang et al.~\cite{yang2018graph} initially only detect and classify objects, but later use a relation-proposal network to predict relationships based on object classes. Yang et al.~\cite{yang2018graph} uses the predicted classes of the objects to hypothesize possible relationships in the relation-proposal network.

Davis et al. \cite{davis2019deep} detects text lines using a Fully Convolutional Network (FCN). They also use dilated, non-squared kernels in the FCN text detector to improve the detection accuracy for long text lines. They use a convolutional classifier network to predict which potential relationships are correct using a context window around the relationship. All pairs of bounding boxes whose edges are within line-of-sight of each other, and are not too far away from each other, are consider candidates.

\section{Implementation and results}
\subsection{Training configurations}
\label{training-config}
As there are many good out-of-the-box text detectors that have been pre-trained on very large datasets, training a new one would not be necessary. Furthermore, it is not feasible to train the full end-to-end model due to limitation on computational power, the key-value detection network will be trained separately using text segmentation map derived from the revised annotation. This is still adequate to prove that key value components can be determined using only spatial information.

\textbf{Data} Out of the 149 images in the training subset, 99 of them are used for training the model and the remaining 50 are used as the validation set. The network is trained with single image input (i.e., batch size = 1).

\textbf{Input} The input of the key-value detection network has two channels, one for the text mask and one for the document image in greyscale.

\textbf{Architecture} We use an U-Net \cite{ronneberger2015unet} based network that has \([16, 32, 64, 128, 256, 128, 64, 32, 16]\) filters for the key-value detector. It uses \(3 \times 3\) convolutional filters for every convolution layers. Following a convolutional layer are the Rectified Linear Unit (ReLU) activation function and a batch normalization layer. Furthermore, instead of transposed convolution \cite{dumoulin2016guide}, resize-convolution (nearest-neighbour interpolation up-sampling followed by normal convolution) was used as the method of choice for up-sampling the feature map. We also use the same network but with normal convolution replaced by CI-Deformable convolution \cite{cideform}, which helps the network more flexible at spatial modeling.

\textbf{Loss Function} A combination of dice loss and categorical cross-entropy loss. The dice loss is only calculated on the first three layers, ignoring the channel corresponding to the background class. For the categorical cross-entropy loss, a weighted version is used. A set of weights proportional to [1.0, 1.0, 1.0, 0.3], which is normalized so that the weights sum up to 1.0, is applied during cross-entropy calculation. This effectively makes the weight of the background be one-third of other classes. Finally, the final loss value is a weighted sum of the dice loss value and the cross-entropy loss value:
\begin{equation}
    loss_{final} = 4 \times loss_{dice} + 0.5 + 0.5 \times loss_{cross-entropy}
    \label{eq:loss_func}
\end{equation}

\textbf{Metric} The key-value detection problem is essentially a segmentation task, so the mean IoU score is used to assess the performance. The IoU value is calculated per each channel (i.e. each class) in the output mask and then averaged to get the mean IoU (mIoU) value. 

\textbf{Other configurations} Kernel weights are initialized with He Normal Initialization [12] and are normalized using L2 normalization with a factor of 0.01. All convolutional layers use “same” padding and kernels of size \(3 \times 3\). During training, Adam optimizer with learning rate = 0.0001 is used. The model is trained for 200 epochs but with early stopping procedure in place, also the weights with the best result will be restored upon termination of training.

\subsection{Results}
As mention earlier, the network takes both the text mask and the document image as input. It is important to point out that using the input document image together with the text mask yields noteworthy better results than using only the text mask. To verify the effect of such modification, an experiment is conducted where a smaller version of the network described in \ref{training-config} (which starts with only 4 kernels in the first convolution block) is trained on each scenario. According to the results shown in \ref{tab:with-vs-without-text-mask}, even though the network is rather small, it still clearly proves that including the input image in the input provides the network with valuable visual information that improves the result by a large margin.

\begin{table}
    \caption{Comparison between using the text mask with and without the document image as input}
    \centering
    \begin{tabular}{lcc}
        \toprule
        \cmidrule(r){1-2}
        Input       & Text mask only    & Text mask + image \\
        \midrule
        Mean IoU    & 0.55              & \textbf{0.69}    \\
        \bottomrule
    \end{tabular}
    \label{tab:with-vs-without-text-mask}
\end{table}

For the main experiment, the result is shown in \ref{tab:unet-vs-cideform-unet}. Since the last channel (corresponding to the background class) is fairly easy to learn, not taking it into account during mIoU calculation would better reflect the performance of the network, but we still provide the result where the background is included for reference. The results show that using only visual information, it is possible for the network to effectively learn to identify “key” and “value” entities in document images. Moreover, the network converges faster and achieves better results when using CI-Deform Conv \cite{cideform} suggests that better spatial modeling capability corresponds to better performance at this task. 

\begin{table}
    \caption{Key and value detection using U-Net \cite{ronneberger2015unet} and U-Net with CI-Deformable convolution \cite{cideform}}
    \centering
    \begin{tabular}{ccc}
        \toprule
        \cmidrule(r){1-2}
            & U-Net    
            & \begin{tabular}{c}U-Net \\ with CI-Deform Conv\end{tabular} \\
        \midrule
            Epochs
            & 205
            & 183 \\
            
            \hline
            Mean IoU
            & 0.79 
            & \textbf{0.80}    \\
            
            \hline
            \begin{tabular}{c}Mean IoU \\ (without background)\end{tabular}
            & 0.72 
            & \textbf{0.73} \\
        \bottomrule
    \end{tabular}
    \label{tab:unet-vs-cideform-unet}
\end{table}

\section{Conclusion}
In this report, we have described some labeling issues in FUNSD, one of the limited available dataset for key-value detection problem in document images.  We have made some revision to the dataset. The revised FUNSD is expected to be a promising dataset to automatically recognize the semantic relationships (key-value relations) in visual data. We also reported our implementation of for key-value detection on FUNSD using a UNet model as baseline results and an improved UNet model with Channel-Invariant Deformable Convolution.

\bibliographystyle{unsrt}  
\bibliography{references}  

\begin{thebibliography}{10}

\bibitem{davis2019deep}
Brian Davis, Bryan Morse, Scott Cohen, Brian Price, and Chris Tensmeyer.
\newblock Deep visual template-free form parsing.
\newblock In {\em 2019 International Conference on Document Analysis and
  Recognition (ICDAR)}, pages 134--141. IEEE, 2019.

\bibitem{jaume2019funsd}
Guillaume Jaume, Hazim~Kemal Ekenel, and Jean-Philippe Thiran.
\newblock Funsd: A dataset for form understanding in noisy scanned documents.
\newblock In {\em 2019 International Conference on Document Analysis and
  Recognition Workshops (ICDARW)}, volume~2, pages 1--6. IEEE, 2019.

\bibitem{harley2015evaluation}
Adam~W Harley, Alex Ufkes, and Konstantinos~G Derpanis.
\newblock Evaluation of deep convolutional nets for document image
  classification and retrieval.
\newblock In {\em 2015 13th International Conference on Document Analysis and
  Recognition (ICDAR)}, pages 991--995. IEEE, 2015.

\bibitem{barrett2004digital}
William Barrett, Luke Hutchison, Dallan Quass, Heath Nielson, and Douglas
  Kennard.
\newblock Digital mountain: From granite archive to global access.
\newblock In {\em International Workshop on Document Image Analysis for
  Libraries}, pages 104--121, 2004.

\bibitem{butt2012information}
Muheet~Ahmed Butt.
\newblock Information extraction from pre-preprinted documents.
\newblock {\em International Journal of Computers and Distributed System},
  2(1):88--93, 2012.

\bibitem{hammami2015one}
Maroua Hammami, Pierre H{\'e}roux, S{\'e}bastien Adam, and Vincent~Poulain
  d'Andecy.
\newblock One-shot field spotting on colored forms using subgraph isomorphism.
\newblock In {\em International Conference on Document Analysis and Recognition
  (ICDAR)}, 2015.

\bibitem{zhou2016irmp}
Jun Zhou, Han Yu, Cheng Xie, Hongming Cai, and Lihong Jiang.
\newblock i{RMP}: From printed forms to relational data model.
\newblock In {\em IEEE HPCC/SmartCity/DSS}, pages 1394--1401, 2016.

\bibitem{hirayama2011development}
Junichi Hirayama, Hiroshi Shinjo, Toshikazu Takahashi, and Takeshi Nagasaki.
\newblock Development of template-free form recognition system.
\newblock In {\em International Conference on Document Analysis and Recognition
  (ICDAR)}, pages 237--241, 2011.

\bibitem{zhang2017relationship}
Ji~Zhang, Mohamed Elhoseiny, Scott Cohen, Walter Chang, and Ahmed Elgammal.
\newblock Relationship proposal networks.
\newblock In {\em IEEE Computer Vision and Pattern Recognition (CVPR)},
  volume~1, page~2, 2017.

\bibitem{yang2018graph}
Jianwei Yang, Jiasen Lu, Stefan Lee, Dhruv Batra, and Devi Parikh.
\newblock Graph {R-CNN} for scene graph generation.
\newblock In {\em European Conference on Computer Vision (ECCV)}, pages
  670--685, 2018.

\bibitem{ronneberger2015unet}
Olaf Ronneberger, Philipp Fischer, and Thomas Brox.
\newblock U-net: Convolutional networks for biomedical image segmentation.
\newblock In {\em International Conference on Medical image computing and
  computer-assisted intervention}, pages 234--241. Springer, 2015.

\bibitem{dumoulin2016guide}
Vincent Dumoulin and Francesco Visin.
\newblock A guide to convolution arithmetic for deep learning.
\newblock {\em arXiv preprint arXiv:1603.07285}, 2016.

\bibitem{cideform}
Hieu~M. Vu and Thi Ngoc~Diep Nguyen.
\newblock Enhancing convolutional neural networks with channel invariant
  deformable convolution.
\newblock 2020.

\end{thebibliography}

\end{document}